\documentclass[10pt,letterpaper]{article}
\usepackage[top=1.0in,left=1.5in,footskip=1.0in]{geometry}

\usepackage{amsmath,amssymb}

\usepackage{changepage}

\usepackage[utf8x]{inputenc}

\usepackage{textcomp,marvosym}

\usepackage{cite}

\usepackage{nameref,hyperref}

\usepackage{microtype}
\DisableLigatures[f]{encoding = *, family = * }

\usepackage[table,dvipsnames]{xcolor}

\usepackage{array}

\newcolumntype{+}{!{\vrule width 2pt}}

\newlength\savedwidth

\usepackage{setspace} 
\raggedright
\usepackage[aboveskip=1pt,labelfont=bf,labelsep=period,justification=justified,singlelinecheck=off]{caption}

\makeatletter
\renewcommand{\@biblabel}[1]{\quad#1.}
\makeatother

\usepackage{lastpage,fancyhdr,graphicx}
\usepackage{epstopdf}
\usepackage{verbatim}

\begin{document}
\vspace*{0.2in}
\begin{flushleft}
{\Large
\textbf\newline{In-home and remote use of robotic body surrogates by people with profound motor deficits} 
}
\newline
\\
Phillip M. Grice\textsuperscript{1,*},
Charles C. Kemp\textsuperscript{1} \\
\bigskip
\textbf{1} Department of Biomedical Engineering, Georgia Institute of Technology, Atlanta, GA, USA
\\
\bigskip

% Use the asterisk to denote corresponding authorship and provide email address in note below.
* phillip.grice@gatech.edu (PG)
\pagebreak
\end{flushleft}
\section*{Abstract}
By controlling robots comparable to the human body, people with profound motor deficits could potentially perform a variety of physical tasks for themselves, improving their quality of life. The extent to which this is achievable has been unclear due to the lack of suitable interfaces by which to control robotic body surrogates and a dearth of studies involving substantial numbers of people with profound motor deficits. We developed a novel, web-based augmented reality interface that enables people with profound motor deficits to remotely control a PR2 mobile manipulator from Willow Garage, which is a human-scale, wheeled robot with two arms. We then conducted two studies to investigate the use of robotic body surrogates. In the first study, 15 novice users with profound motor deficits from across the United States controlled a PR2 in Atlanta, GA to perform a modified Action Research Arm Test (ARAT) and a simulated self-care task. Participants achieved clinically meaningful improvements on the ARAT and 12 of 15  participants (80\%) successfully completed the simulated self-care task. Participants agreed that the robotic system was easy to use, was useful, and would provide a meaningful improvement in their lives. In the second study, one expert user with profound motor deficits had free use of a PR2 in his home for seven days. He performed a variety of self-care and household tasks, and also used the robot in novel ways. Taking both studies together, our results suggest that people with profound motor deficits can improve their quality of life using robotic body surrogates, and that they can gain benefit with only low-level robot autonomy and without invasive interfaces. However, methods to reduce the rate of errors and increase operational speed merit further investigation.

\section*{Introduction}
Individuals with profound motor deficits currently require assistance from human caregivers to complete many physical self-care tasks. This care can create financial challenges in providing for professional caregivers, and can place both physical and emotional burdens on informal caregivers. Assistive robots that enable people with profound motor deficits to perform tasks for themselves could be beneficial. For example, the ability to care for oneself (self-care self-efficacy) correlates with improved quality of life and decreased depression in stroke patients\cite{RobinsonSmith}, and reducing care burden may improve caregiver mortality rates\cite{Schulz1999}.

Robotic body surrogates have the potential to be highly versatile assistive devices for people with profound motor deficits. For this paper, we define a robotic body surrogate to be a human-operated, remote-controlled robot that can navigate and manipulate within an environment in a manner comparable to an able-bodied human. In this way, a robotic body surrogate can serve as a substitute for the human operator being physically present in the environment. A robotic body surrogate can also have physical capabilities that the human operator does not posses, and thereby serve as an assistive device for people with motor impairments. 

\subsection*{Controlling Robotic Body Surrogates is a Key Challenge}

While the versatility of robotic body surrogates could enable them to assist with a wide variety of tasks, they are complex devices with many sensors and degrees of freedom. There are many commercially available human-scale robots with wheels and one or two arms (i.e., mobile manipulators) that could potentially serve as robotic body surrogates \cite{HSR, Fetch, Tiago, KinovaMOVO, RB-1}. Additionally, full humanoid robots with arms and legs have been produced by companies and researchers and are common in laboratories \cite{SpenkoDRC, SHR}. The studies in this paper use a PR2 robot from Willow Garage that is comparable to other commercially available mobile manipulators. The PR2 is a human-scale robot with an omnidirectional wheeled base, a torso that translates vertically, two arms with grippers, a pan/tilt head with cameras, and various other sensors, such as tactile sensors.

Due to this complexity, enabling a single person to effectively control a robotic body surrogate is a critical challenge. In the DARPA Robotics Challenge (DRC), teams used an average of two or more able-bodied experts using six or more video displays to control human-scale robots that navigated and manipulated in a manner comparable to able-bodied humans \cite{DRC_HRI_analysis}. The DRC was a prominent international competition involving well-regarded teams from around the world, yet the interfaces to control the robots were generally cumbersome and would be ill-suited for a single operator.  

The challenge of single person control becomes even greater when the person has profound motor deficits. In this paper, we consider a person who scores nine or fewer points on the Action Research Arm Test (ARAT) with both upper limbs to have profound motor deficits. This corresponds with a person having limited voluntary motion of his or her upper limbs, typically such that the person is unable to lift either hand against gravity. Profound motor deficits limit how users can provide input to computer systems \cite{Mankoff2002}, and variations in peoples' impairments and preferences make it difficult to design a single broadly accessible input \cite{Bien04}.

\subsection*{Approaches to the Control of Robotic Body Surrogates}

One approach to overcoming the challenge of interfacing a human with a robotic body surrogate has been to give the robot autonomous capabilities \cite{DRC_HRI_analysis, Murphy2015, Marion2017, SpenkoDRC}. This has the benefit of reducing the responsibilities of the human operator, which can reduce cognitive load, reduce errors, and increase efficiency. However, to date, autonomous capabilities have tended to be narrow in scope, such as a system for self-care tasks around the head \cite{Hawkins2014}, or unproven in broader contexts, such as efforts to enable semi-autonomous grasping and placement of objects \cite{Ciocarlie2012, RFHRAM2013, Tidoni2017}.

Another approach to overcoming this challenge has been through interfaces that more directly connect the human brain to the robot via brain-computer interfaces (BCIs). This approach has the potential to result in higher-bandwidth, lower-latency, and more intuitive interfaces to robotic body surrogates. To date, however, these efforts have focused on control of less complex systems than robotic body surrogates and have significant practical challenges. For example, cortical BCIs \cite{Hochberg2006, Hochberg2012, Collinger2013, Ajiboye2017, Wodlinger2014, Pandarinath2017} show promise for the control of robot arms, but the interfaces are highly invasive and remain an immature technology\cite{Ajemian2017}. As another example, researchers have enabled a user in an MRI machine to remotely control a humanoid robot \cite{Cohen2012}, but MRI machines are currently impractical for home use. 

Our approach is to provide an augmented-reality (AR) interface running in a standard web browser with only low-level robot autonomy. The AR interface uses state-of-the-art visualization to present the robot's sensor information and options for controlling the robot in a way that people with profound motor deficits have found easy to use. In order to meet the needs of users with profound motor deficits, we used participatory design from the outset, involving people with disabilities in the iterative development of the interface \cite{Schuler1993, Sanders2003}. The standard web browser enables people with profound motor deficits to use the same methods they already use to access the Internet to control the robot. Many commercially-available assistive input devices, such as head trackers, eye gaze trackers, or voice controls, can provide single-button mouse-type input to a web browser. Identifying the most appropriate assistive input device for an individual is a common challenge in assistive technology, and is based upon each individual’s specific deficits.
Due to the great value associated with accessing the Internet, people who have profound motor deficits often learn to use a web browser, either via a commercially-available assistive input device or custom accommodations \cite{Simpson2010}. By limiting the robot's autonomy to low-level operations, such as tactile sensor driven grasping and moving an arm with respect to inverse kinematics to achieve end effector poses, the robot performs consistently across diverse situations allowing the user to attempt to use the robot in diverse and novel ways. 

\subsection*{Can People with Profound Motor Deficits Benefit?}

The extent to which people with profound motor deficits can benefit from robotic body surrogates has been unclear. Relevant published work has primarily involved only a small number of participants, often a single participant, making the extent to which the results would generalize across other people with profound motor deficits uncertain \cite{RFHRAM2013, Park2017, Hawkins2014, Tsui2015, Wang13, Soekadar2016, Bien04}. The great diversity of deficits and causes of deficits increases this uncertainty. In addition, there has been a lack of evaluations based upon clinical assessments. 

To address these limitations, we conducted two studies to investigate the use of robotic body surrogates. In the first study (Study 1), 15 novice users with profound motor deficits from across the United States controlled a PR2 in Atlanta, GA to perform a modified ARAT and a simulated self-care task. For this study, our experimental design and web-based interface greatly increased the size of the population from which we could recruit, since participants could be located across the country and did not need to travel to the laboratory. Many participants operated the robot from their homes. This made it feasible for us to meet our recruiting goals as determined by a statistical power analysis. 

We designed Study 1 to characterize the benefit or lack of benefit provided by a robotic body surrogate in terms of an established clinical assessment (i.e., the ARAT) and to establish a performance baseline for this assessment. In addition, we used a simulated self-care task to assess performance with respect to a task that required both navigation and manipulation. We also measured perceptions of the system's ease of use, usefulness, and potential to meaningfully improve people's lives.

We designed the second study (Study 2) to characterize the use of a robotic body surrogate by an expert user with profound motor deficits in his home over an extended period of time. This study complements Study 1 by providing insights into how robotic body surrogates might be used on a daily basis in a real home, whereas Study 1 examined relatively brief use by novices in a controlled laboratory setting. Robotic systems that have only been tested in laboratory settings often fail when tested under real-world conditions, which can lead to misinterpretation of results \cite{Alaiad2014, Sung2010}. 

\subsection*{Overview}

The next section, Methods, presents descriptions of the robot hardware, the user interface, Study 1, and Study 2. The subsequent section, Results, presents the results obtained from Study 1 and Study 2. Then, we provide a section, Discussion, with commentary on various aspects of the research, including limitations of the robotic system and implications for robotic body surrogates. Finally, we provide a brief Conclusion section that summarizes the main outcomes of the work. 

\section*{Methods}

In this section, we present the methods we used for this research, starting with a brief description of the robot hardware followed by a detailed presentation of the interface and descriptions of Study 1 and Study 2. 

\subsection*{Robot Hardware}
\begin{figure}[h!]
\centering
\includegraphics[width=\textwidth]{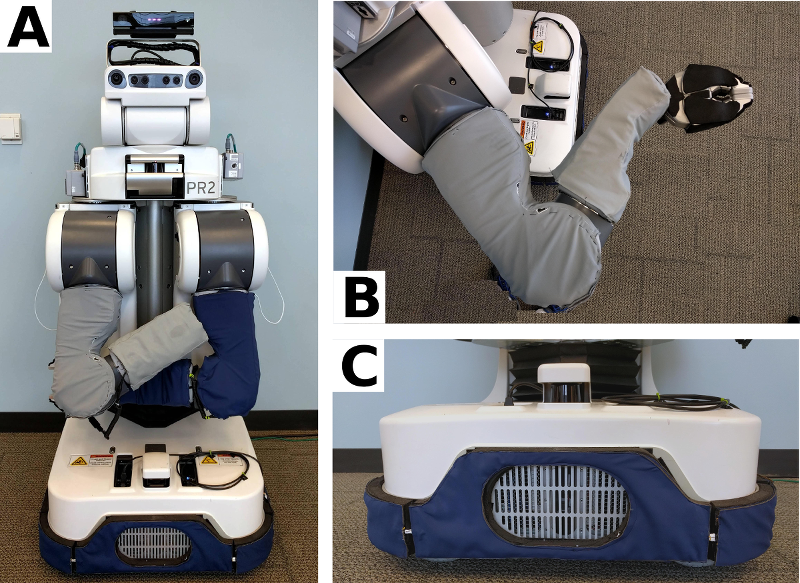}
\caption{{\bf The robotic body surrogate (Willow Garage PR2).}
(A) The PR2 robot.
(B) One of the robot's seven DoF arms, including the tactile-sensing fabric skin (gray) and foam padding (black) on the metallic gripper.
(C) The base of the robot, including tactile-sensing fabric skin (blue), placed atop foam padding.
}
\label{fig:PR2_parts}
\end{figure}

For Study 1 and Study 2, we used a PR2 robot running Ubuntu 14.04 and the open-source Robot Operating System (ROS) Indigo Igloo.
We modified a standard PR2 by adding dense foam padding around the metal grippers and the mobile base of the robot to reduce potential negative consequences of contact. We also added fabric-based tactile sensing skin\cite{Jain2013, Grice2013} around the mobile base and to both upper arms and forearms (Fig. \ref{fig:PR2_parts}B,C). 
We adjusted the control gains on the PR2's backdrivable arms to increase their compliance, and used low-velocity arm motions for additional safety.
We added a Microsoft Kinect sensor to the robot's head, which provides a 1920x1080 pixel RGB color video and a 512x424 pixel depth video.
When used with our interface, the robot has 20 controllable degrees of freedom.
Instructions for recreating the pressure sensitive fabric skin are available at \newline\url{http://pwp.gatech.edu/hrl/manipulation-with-whole-arm-tactile-sensing/}.
The source code for our system and interface is available in \nameref{source_code}. 
We are aware of at least one research group that has successfully run the system described (without the custom tactile sensors) on another PR2.

\subsection*{A Novel Web-based Augmented Reality Interface}

We developed a novel,  web-based augmented reality (AR) interface that maps single-button mouse-type input to motions of a PR2 robot (Fig. \ref{fig:PR2_parts}). Users access the interface using a standard web browser\cite{RWT}, simplifying access and enabling control of a robot away from the user, such as in another room or across the country.

\subsubsection*{Participatory Design}

We developed the interface through an iterative, collaborative participatory research process \cite{Cornwall1995, Viswanathan2004, Cargo2008, Schuler1993, Sanders2003} with Henry Evans, an individual with severe quadriplegia from a brain stem stroke, and Jane Evans, his wife and primary caregiver.

Researchers have previously noted the importance of including all interested parties, including caregivers, in the development of assistive technology \cite{Kintsch2002, Wilkinson2014}.
While some researchers have included users in the design process \cite{Wagner1999, Green2000}, they more often focus on the technical challenges during development, and only receive feedback from users during evaluation.
We have worked with Henry, Jane, and others to develop assistive robotic technologies since 2011 \cite{RFHRAM2013,Hawkins2014}.
Lessons learned from these efforts led to the development of the system reported here.

During the development of this system, we met with Henry and Jane Evans weekly using video-conferencing software.
We conducted remote evaluations of new functionality approximately monthly to explore design ideas and receive user feedback.
Approximately two times per year, researchers traveled to conduct in-person evaluations in the Evans's home, culminating in the seven-day evaluation described below in Study 2.
We used the insights gained through these discussions and evaluations to identify both user needs and system improvements to enable effective use by individuals from the target population.
We have previously described their involvement and some aspects of the web interface design \cite{Grice2016design}, and present the final hardware and software system used in the subsequent evaluations here.

\subsubsection*{Single-button Mouse-type Input}

\begin{figure}[t]
\centering
\includegraphics[width=0.9\textwidth]{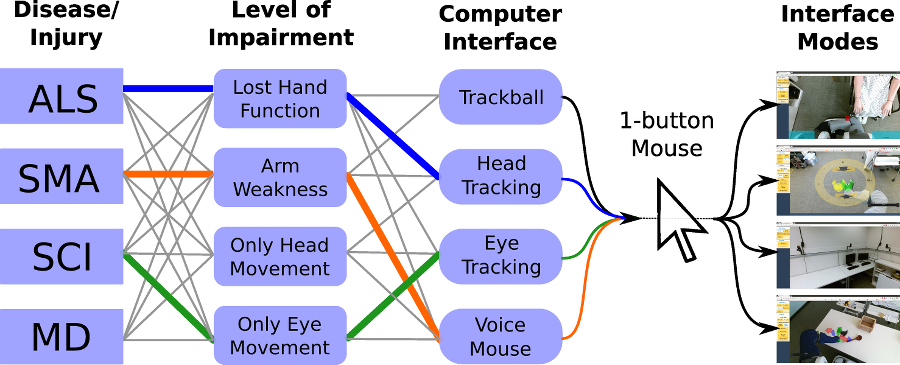}
\caption{{\bf Enabling system operation through single-button mouse-type input simplifies design and provides broad accessibility.}
Individuals with diverse disease or injury conditions likely have diverse and possibly changing levels of impairment. These individuals may choose to use a variety of commercially-available, off-the-shelf input devices that enable single-button mouse-type input, which can be used to operate our robotic body surrogate.
The many possible combinations of disease/injury, impairment, and usable computer interface are connected here by gray lines. 
These devices make our system accessible across a range of sources of impairment and personal preferences.
Also, system developers only need to support a single mode of interaction, reducing development and support effort.
Examples: (Blue line) An individual with ALS may have limited hand function and choose to use a head-tracking mouse; (Orange line) An individual with spinal muscular atrophy (SMA) may experience upper-extremity weakness, and prefer the use of a voice-controlled mouse; 
(Green line) An individual with a spinal cord injury (SCI) may only retain voluntary eye movement, and use an eye-gaze based mouse.  All three of these individuals can operate our system without modification, making it accessible across types and sources of motor impairment.}
\label{fig:hci_devices}
\end{figure}

Because many commercially-available assistive input devices can provide single-button mouse-type input to a web browser, designing assistive technology for use with standard mouse-type input simplifies system development, reduces the need to develop specialized interfaces \cite{Bien04}, and promotes accessibility across medical conditions, impairments, and preferences (Fig. \ref{fig:hci_devices}).
Brain-computer interfaces and other novel assistive input devices can also provide this type of input, making them a complementary technology\cite{Hochberg2006, Pandarinath2017, Townsend2010, Jain2015}.
Additionally, while designed for use by individuals with motor impairments, this access method is also applicable to non-motor-impaired operators, and so is representative of universal design \cite{UniversalDesign}.

\subsubsection*{Video-centric First-person Perspective}

For the interface we developed, the user moves the cursor across a video-centric display \cite{Tsui2015} of a live video stream from the robot's head-mounted camera, while AR interface elements overlaid on the video show sensor data \cite{Zalud2006}, provide direct manipulation controls \cite{DirectManipulation}, and convey how the robot will move when commanded with the mouse button \cite{Chou01}.
The video-centric display uses the camera view from the robot's head to provide the user with feedback on performed actions and context for both planned actions and sensor data.

Other robotic manipulation interfaces, such as ROS RViz, often rely on 3D-rendered displays of the robot and surrounding environment, and require the user to manipulate both the robot and the virtual camera view.
In a prior study, we found that even able-bodied users with experience in virtual 3D modeling had difficulty controlling a robot effectively using this type of interface, despite training and a brief practice session \cite{Grice2013}.
After using a similar interface, \cite{Leeper2012} notes that “the operator’s comfort with a general 3D GUI and related operations such as positioning a virtual camera proved to be very important” for effective task performance.
In contrast, by restricting the user's perspective to the view from the robot's head, our interface eliminates the requirement for the user to position and orient a virtual camera in 3D space, and avoids possible confusion caused by multiple simultaneous views.
This consistent first-person perspective also aids the user in assuming the role of the robot, as this is similar to the perspective from which individuals experience their daily lives.

\subsubsection*{Additional Sensory Feedback}

\begin{figure}[h!]
\centering
\includegraphics[width=\textwidth]{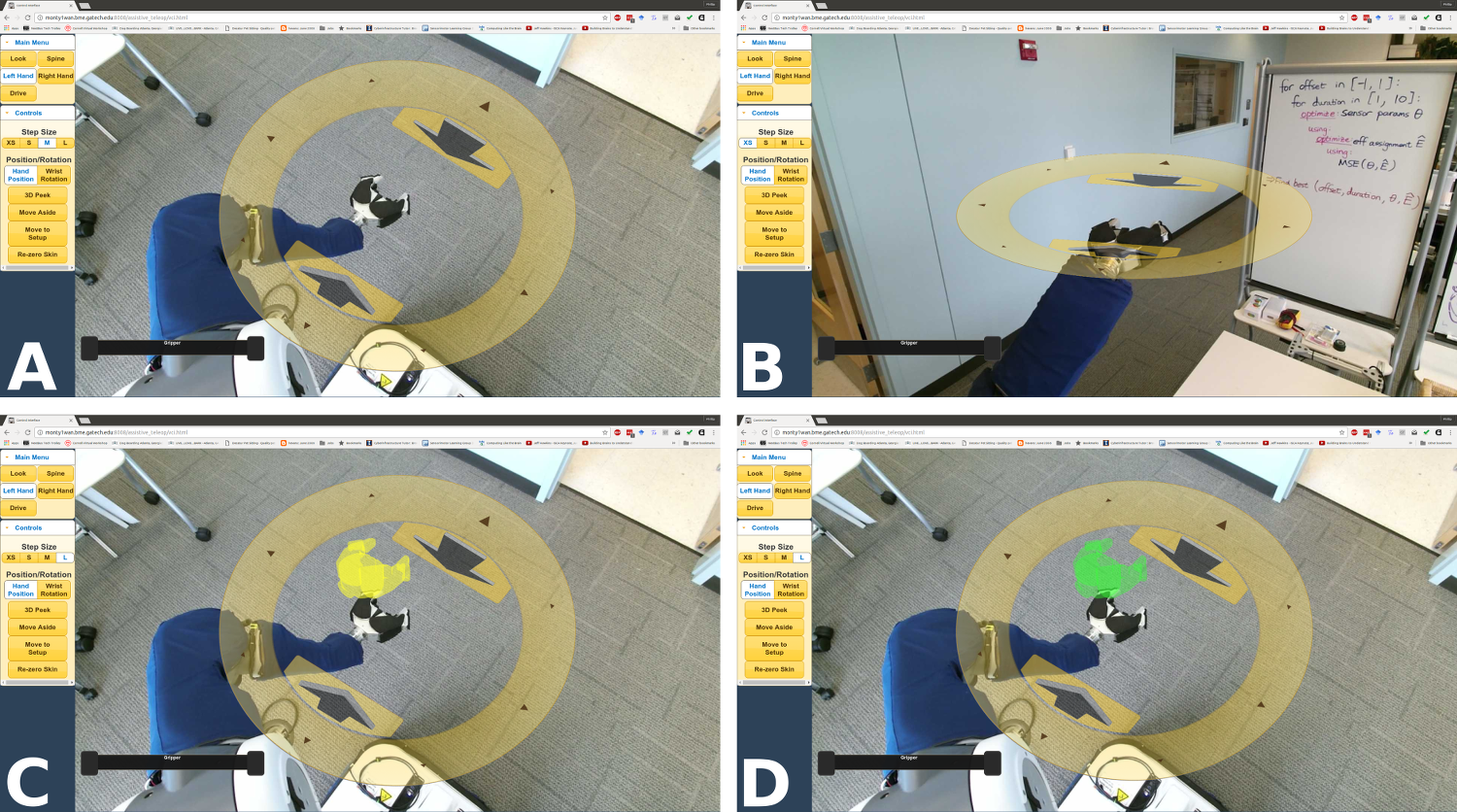}
\caption{{\bf The end-effector position control ring augmented reality interface with virtual preview (yellow) and goal (green) gripper displays.} 
(A) The control ring's rotation remains aligned with the robot's body.
(B) The control ring appears parallel to the floor to convey vertical height.
(C) A yellow virtual gripper `previews' commands by displaying the pose the gripper will attempt to reach if commanded.
(D) A green virtual gripper displays the gripper's current goal, and disappears once it reaches this goal.}
\label{fig:position_interface}
\end{figure}

\begin{figure}[h!]
\centering
\includegraphics[width=\textwidth]{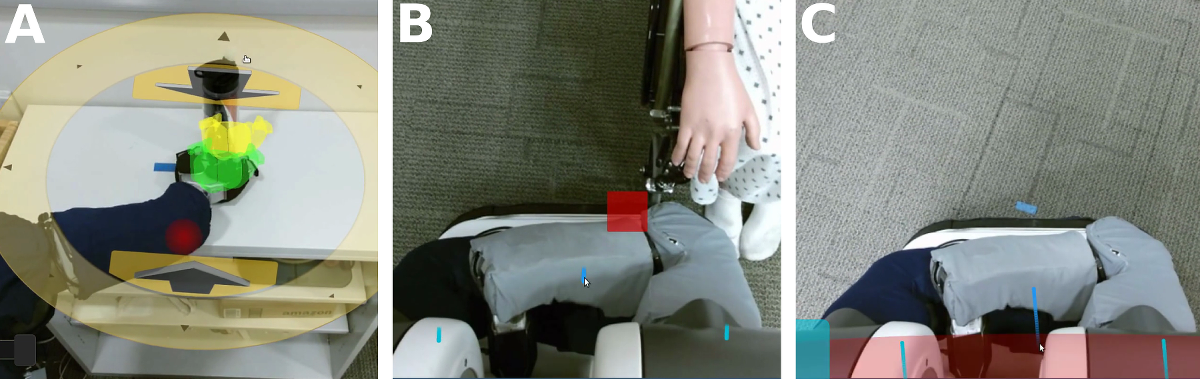}
\caption{{\bf Contact displays overlaid on the video interface based on data from the fabric-based tactile sensors.} 
(A) Contact on the forearm against the table edge.
(B) Contact between the robot's base and the wheelchair.
(C) Contact with the robot's base behind the current field of view.}
\label{fig:contact_displays}
\end{figure}

In addition to the video stream, the interface displays other sensor data from the robot in an integrated manner. For example, the interface uses joint angle sensing and a kinematic model to display interface elements around the robot's gripper in the video stream (Fig. \ref{fig:position_interface}). As another example, the interface displays the output of the robot's pressure sensitive skin. If the fabric-based tactile sensors on the arms or base detect contact, a red dot or square, respectively, appears in the camera view at the location of contact (Fig. \ref{fig:contact_displays}A,B). 
If contact occurs outside of the camera view, the nearest edge or corner of the screen flashes red (Fig. \ref{fig:contact_displays}C).

\begin{figure}[h!]
\centering
\includegraphics[width=\textwidth]{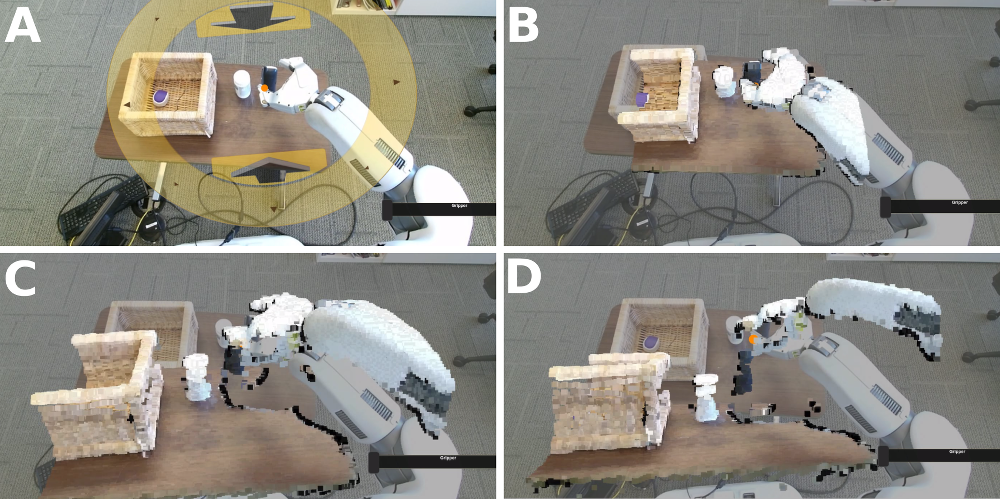}
\caption{{\bf The 3D Peek feature showing the 3D point cloud over the live camera feed, rotated to provide depth perception.} 
(A) The view before the 3D Peek.
(B) $\approx0.1s$ into the 3D Peek.
(C) $\approx0.3s$ into the 3D Peek.
(D) 3D Peek view (holds for $2.8s$).}
\label{fig:3d_peek}
\end{figure}

To aid depth perception, the interface provides a ``3D Peek" feature, which overlays a down-sampled, RGB-D point-cloud of the volume around the gripper onto the video, using data from the Kinect sensor. 
This simulated view then virtually rotates, as if the camera lowers from the robot's head to the height of the gripper, allowing a simulated view from this height (Fig. \ref{fig:3d_peek}). ``3D Peek" is available in the ``Right/Left Hand'' modes described later.

\subsubsection*{Modal Control}

\begin{figure}[h!]
\centering
\includegraphics[width=\textwidth]{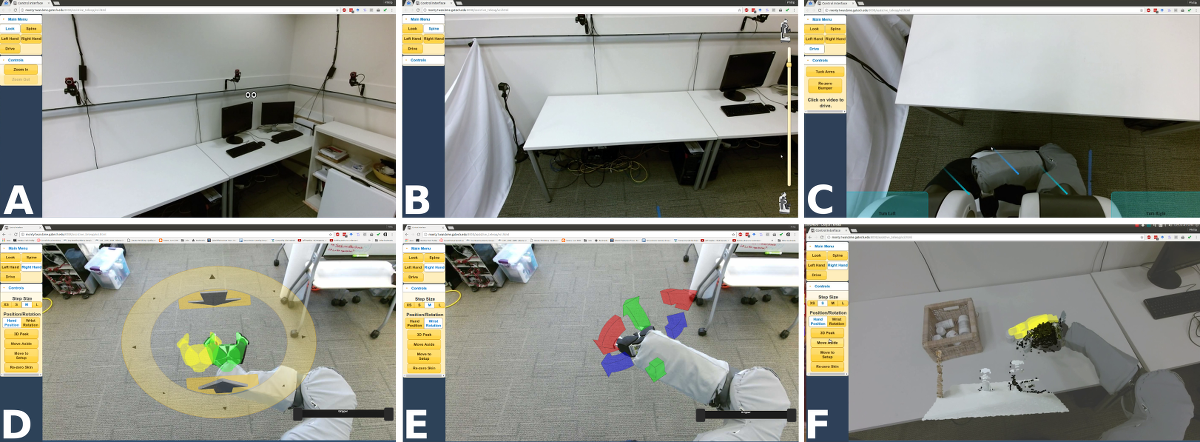}
\caption{{\bf The interface used to operate the robotic body surrogate.} 
(A) `Looking' mode.
(B) `Spine' mode.
(C) `Driving' mode.
(D) `Hand position' mode.
(E) `Hand rotation' mode.
(F) `3D Peek' depth display.}
\label{fig:interfaces}
\end{figure}

The interface uses modal control, where the same input has a different output depending upon the active mode (Fig. \ref{fig:interfaces}).
Modal control introduces the opportunity for mode errors, where the correct command, given in the wrong mode, produces an undesired result \cite{NormanDOET}, and can create mode-switching delays \cite{Herlant2016}.
However, dividing control of the robot's degrees of freedom across multiple control modes allows all of the degrees of freedom to be controlled using only a mouse-type input, and also reduces visual clutter and the number of immediately available input options. 

We reduce mode-switching delays by making all primary modes selectable from a top-level menu on the left of the screen, and by allowing the server-side control components to run concurrently, which means that mode switching only requires client-side interface changes. To help avoid mode errors, each mode uses visually distinct AR elements to convey the current mapping from the mouse cursor and single button to robot motions (Fig. \ref{fig:interfaces}, \nameref{S1_Video}), and to display relevant sensor data in the appropriate context in the camera view.

The robot executes step-wise motions for all modes except driving, but the web browser renders the AR elements on the user's machine. This allows the interface to provide responsive, real-time feedback to the user's input at all times, while the robot is able to perform at least small actions independently, without the jitter and lag often associated with streaming commands over unreliable networks.

The operation modes within the interface are as follows.
\begin{description}
\item [``Looking'' mode] displays the mouse cursor as a pair of eyeballs, and the robot looks toward any point where the user clicks on the video.
\item [``Driving'' mode] allows users to drive the robot in any direction without rotating, or to rotate the robot in-place in either direction. 
The robot drives toward the location on the ground indicated by the cursor over the video when the user holds down the mouse button, and three overlaid traces show the selected movement direction, updating in real time. 
``Turn Left'' and ``Turn Right'' buttons over the bottom corners of the camera view turn the robot in place.
\item [``Spine'' mode] displays a vertical slider over the right edge of the image.
The slider handle indicates the relative height of the robot's spine, and moving the handle raises or lowers the spine accordingly.  
These direct manipulation features use the context provided by the video feed to allow the user to specify their commands with respect to the world, rather than the robot, simplifying operation.
\item [``Left Hand'' and ``Right Hand'' modes ] allow control of the position (Fig. \ref{fig:position_interface}) and orientation (Fig. \ref{fig:rotation_interface}) of the grippers in separate sub-modes, as well as opening and closing the gripper.
In either mode, the head automatically tracks the robot's fingertips, keeping the gripper centered in the video feed and eliminating the need to switch modes to keep the gripper in the camera view. 
\end{description}

\subsubsection*{Details for the ``Left Hand'' and ``Right Hand'' Modes}

The ``Left Hand'' and ``Right Hand'' modes are essential for manipulation, playing a critical role in both Study 1 and Study 2. These modes also include a number of novel attributes. As such, we now provide a more detailed description.

\paragraph{Gripper Position Sub-mode} To control the gripper's position, the user clicks on a yellow virtual disk displayed around the gripper. This moves the gripper one step on a horizontal plane toward the corresponding 3D point on the virtual disk.
Step sizes can be selected from XS, S, M, and L ($1.5, 4, 11,$ and $25\ cm$, respectively).
The selected step size remains in effect for all movements of the selected hand until adjusted by the user.
Inset up and down arrow buttons move the gripper one step vertically up or down.
The disk tilts to appear co-planar to the floor and rotates so the top points parallel with the robot's base, providing additional situational awareness (Fig. \ref{fig:position_interface}A,B). As a whole, this novel interface simplifies Cartesian motions with respect to the environment. The interface element also provides some of the advantages of direct manipulation interfaces \cite{DirectManipulation}. Clicking a location results in the gripper moving toward both the clicked 2D location in the video and the clicked 3D location in the real world, where the real-world 3D location is defined by the clicked point on the virtual disk. In addition, the rendering of the interface element provides cues to the 3D position of the gripper with respect to the camera, since it is rendered as though it were a horizontal object with a fixed orientation relative to the robot's base. Notably, the interface element does not obscure the user's view of the center of manipulation. 

\begin{figure}[h!]
\centering
\includegraphics[width=\textwidth]{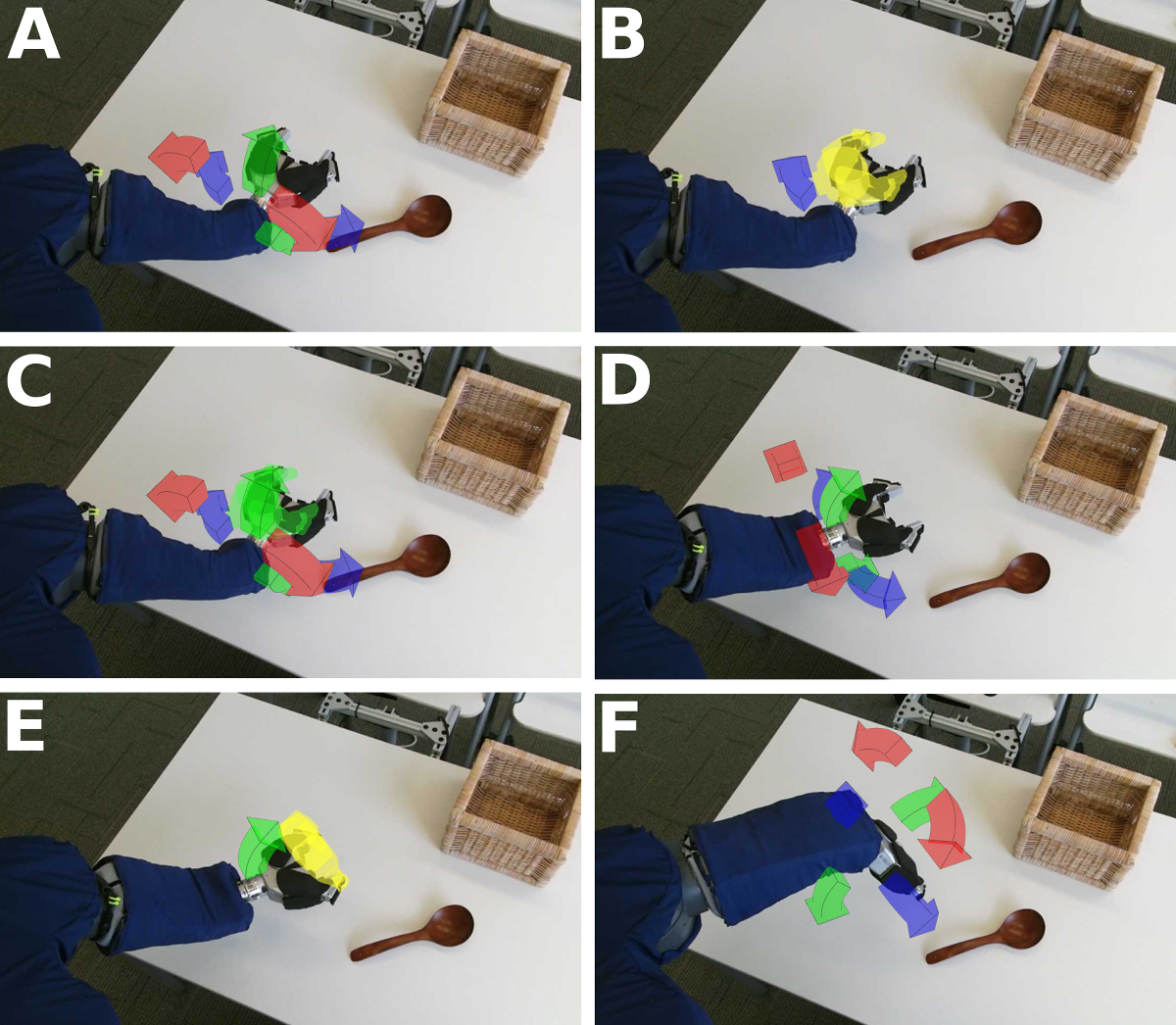}
\caption{{\bf The end-effector orientation control augmented reality interface with virtual preview (yellow) and goal (green) gripper displays.} 
(A) 3D virtual orientation controls around end effector.
(B) Hovering over the blue arrow hides other arrows and shows yellow preview.
(C) After sending a command, a green virtual gripper shows the active goal.
(D) Gripper position after rotating to left from (A).
(E) Hovering over green arrow hides arrows, shows preview.
(F) Gripper position after rotating upward from (E).}
\label{fig:rotation_interface}
\end{figure}

\paragraph{Gripper Orientation Sub-mode} To control the gripper's orientation, six colored, virtual arrows around the robot's wrist move the wrist in the indicated direction when clicked, rotating the gripper about the fingertips (Fig. \ref{fig:rotation_interface}).   
Step sizes can be selected from XS, S, M, and L ($\frac{\pi}{18}, \frac{\pi}{10}, \frac{\pi}{5},$ and $\frac{\pi}{3}\ rad$, respectively). 
These virtual arrows are rendered to always appear in the same location and orientation relative to the gripper as the gripper moves.
This display reduces visual clutter around the gripper \cite{Rosenholtz07}, while providing a consistent interaction, as the gripper will always rotate in the direction indicated by the arrow, unlike alternative interfaces that provide button-pads of directional arrows, where the user must mentally map each command to the current orientation of the gripper.

\paragraph{Grasping} In both the position and orientation sub-modes, the user can open or close the gripper using sliders in the bottom left or right corner of the screen.
While closing, the gripper attempts to grasp items gently but securely using the method described in \cite{Romano2011}.

\paragraph{Gripper Previews} When the cursor hovers over any end effector control, a yellow, semi-transparent, virtual gripper appears where the command being considered would send the gripper, providing a preview \cite{Chou01,Ciocarlie2012}. 
This preview allows users to rapidly evaluate the correctness of their currently selected input to help avoid errors or make adjustments without actually moving the robot and possibly causing unintended interactions (such as knocking over a glass).
When a command is sent, a green, semi-transparent, virtual gripper appears at the goal location and disappears once the goal is reached.

\subsection*{Study 1: Remote Evaluation using a Modified Action Research Arm Test and a Simulated Self-care Task}

Study 1 characterizes the benefit or lack of benefit provided by a robotic body surrogate in terms of a modified version of the ARAT \cite{Lyle1981_ARAT,Yozbatiran2008,Lang2008} and also establishes a performance baseline with respect to this assessment. In addition, Study 1 assesses performance on a simulated self-care task that consists of using the robot to retrieve a water bottle and then bring the tip of the water bottle's straw to the mouth of a medical mannequin.  Study 1 uses questionnaires to measure perceptions of the robotic system. 

Our inclusion criteria required participants at least 18 years of age, fluent in written and spoken English, able to operate a computer mouse or equivalent assistive device, and scoring nine or fewer points on the ARAT with both upper limbs.
We prescreened participants verbally before enrollment.
All participants were compensated \$25.00 USD per hour of participation at the end of the study.
The Georgia Institute of Technology Institutional Review Board (IRB) approved this study under protocol H13046.
We obtained written informed consent from all participants, and all procedures were carried out according to the approved protocol guidelines.

\subsubsection*{Session 1: Demographics, Technology Use, and Unassisted ARAT}

During the first session, we collected demographic information (age, gender, ethnicity, marital status, highest level of education completed, dominant hand, cause of motor deficit, and date of accident/injury/diagnosis). 
We asked participants about their prior use (if any) of robots, video games, and computer aided design software, as well as for details about their computer system, Internet bandwidth, and mouse device. 
We then remotely assessed motor deficit according to the ARAT by asking participants to report their ability to perform each of the ARAT items.
The low ARAT score required for inclusion made this remote assessment feasible.
If scoring for an item was unclear, we used a conservative score estimate (recording the highest possible score/least impairment).
Participants who passed prescreening, but did not meet the criteria for impairment based on the ARAT, were not advanced to the next session. 

\subsubsection*{Session 2: Guided Training and Practice}

Before the second session, we provided a link to a 10-minute tutorial video introducing the robot and the control interface, and a link to a Fitts's law-based pointing test based upon \cite{FittsLawDetails}. This allows estimation of the participant's throughput with their preferred pointing device.
During the second session, participants used the web-based interface to operate a PR2 robot through a guided training session, which introduced all the features of the control interface, and included grasping and placing two plastic bottles from a tabletop. 
Participants then completed a training evaluation task using the robot without guidance.
The task required coordinated use of system features to grasp a box from a nearby shelf and return it to a table in the same room. Participants who failed to complete the training task in under 35 minutes did not proceed further in the study.
We specifically avoided using ARAT tasks as part of this training and practice, so that participants would not learn skills specific to the ARAT. We required at least one overnight period after the training session before proceeding to the third session.

\subsubsection*{Session 3: Modified ARAT with the Robot}
In the third session, participants remotely operated the arm and gripper of the PR2 corresponding to their own dominant arm to complete the ARAT.
The base, spine, and other arm controls were unavailable during this portion of the experiment.
This makes the test more comparable to the ARAT as administered directly to people, but also reduces the system to nine controllable degrees of freedom (pan/tilt head, open/close gripper, and six degree of freedom gripper pose).
We administered the test as closely as possible to published instructions\cite{Yozbatiran2008}, skipping the 10 cm block, flat washer, and ball bearing, which the robot's gripper cannot grasp.
Skipped items were scored as failures (0 points).
We treat all finger combinations aside from Thumb and 1\textsuperscript{st} finger as amputations (0 points), as the robot has only a two-finger gripper.
We also allow up to eight minutes to complete each task, rather than the standard one minute, though we require completion in less than five seconds for full points on each item, per \cite{Yozbatiran2008}.
These considerations result in an expected maximum score using the robotic body surrogate of 22/57 possible points on the ARAT.
After each task, an automated script returned the robot to the setup configuration.
For gross movement items, we positioned the robot near a mannequin in a wheelchair, such that the mannequin's head was centered along the mid-line between the robot's center and the shoulder of the arm being tested, facing perpendicularly to the robot, and pointed in the direction of the arm being tested. 

After completing the ARAT, we asked participants to complete a debriefing questionnaire.
The questionnaire contained the following items, about which we asked participants to rate their agreement using a seven-point scale: 
\begin{enumerate}
\item The robotic system is easy to use for performing manipulation tasks.
\item The robotic system is useful for performing manipulation tasks.
\item I would prefer the robotic system to a human caregiver for manipulation tasks.
\end{enumerate}

We then asked participants to complete the following sentence from the provided list of options: ``Using the robotic system rather than my own arms would make my ability to perform manipulation tasks...'' 1. Much worse, 2. Meaningfully worse, 3. A little worse, but not meaningfully, 4. Neither better nor worse, 5. A little better, but not meaningfully, 6. Meaningfully better, 7. Much better.
We structured this sentence and options to correspond to the literature on identifying minimal clinically important differences\cite{Lang2008}.
Finally, we asked the participant to provide ``any additional comments or feedback about the system.''
We use a 1-tailed Wilcoxon signed rank test to compare ARAT scores using and not using the robot,  and a 1-tailed 1-sample Wilcoxon signed rank test to compare improvement and rating scale responses to comparison values.

\subsubsection*{Session 4: Simulated Self-care Task with the Robot}

In the fourth and final session, participants remotely controlled the robot with all controls, and all 22 degrees of freedom, available to simulate getting themselves a drink.
Participants drove the robot to grasp a water bottle from a shelf, and then brought the tip of a straw in the water bottle to the mouth of a medical mannequin seated in a wheelchair in the same room.
To indicate success, we inserted a small, round neodymium magnet in the center of the mannequin's mouth, behind the rubber skin.
We placed an M4 x 6 mm ferrous socket cap machine screw loosely into the end of the straw during the trial, and declared success when the screw adhered itself to the magnet.
This required the screw in the tip of the straw to be $<1\ cm$ from the center of the mannequin's mouth.
Unlike the constrained ARAT evaluation, this task required the participants to operate the complete robotic system, including combining mobility, altering the height of the spine, and fine manipulation for grasping and reaching in a simulated real-world environment.
After completing the task, we asked participants the same debriefing items as above, replacing `manipulation tasks' with `self-care tasks' in all items.
We also asked ``if you had this robotic system in your home, what tasks would you use the robot for in your daily life?''

\subsubsection*{Enrollment, Demographics, and Computer Access Methods}

We enrolled 37 participants after a brief prescreening process.
14 participants failed to meet the full inclusion criteria after fully evaluating their ARAT score based upon their own physical capabilities, and did not proceed further in the study.
Of the remaining 23 participants, two withdrew before completion citing lack of time, and another withdrew citing personal health. 
One participant with advanced Amyotrophic Lateral Sclerosis (ALS) became ill and passed away before completing participation.
One participant did not have sufficient Internet bandwidth to operate the robot remotely, and two participants failed to schedule beyond the first session before the study was ended.
One participant failed to complete the training evaluation task in the required 35 minutes in two attempts, after receiving 180 minutes of guided training and practice time.
The remaining 15 participants completed the entire study successfully, and all following results derive from these participants.

\begin{figure}[h!]
\centering
\includegraphics[width=0.60\textwidth]{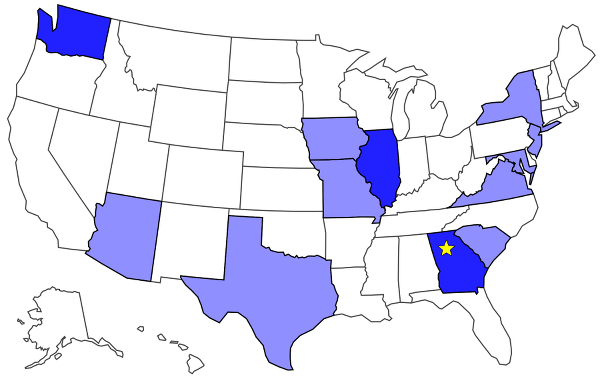}
\caption{{\bf Locations of the 15 participants with profound motor deficits who remotely operated a robotic body surrogate in Atlanta, GA (star) across long distances.} This evaluation used participants' own, existing computer hardware and Internet connections, demonstrating our system's performance with real-world bandwidth and latency constraints. Darker states indicate two participants from that state, lighter states indicate one participant. }
\label{fig:map}
\end{figure}

Participants' motor deficits arose from six distinct sources: Spinal Muscular Atrophy (n=6), Muscular Dystrophy (Duchenne or Becker, n=3), Spinal Cord Injury (n=3), ALS (n=1), Arthrogryposis (n=1), and Dejerine-Sottas disease (n=1).
These 15 participants had an average age of 36.9 $\pm$ 8.7 years, included 8 men, and 13 were right-hand dominant.
Participants identified themselves as Caucasian (n=12), African American (n=2), and Asian (n=1).
Participants had the following levels of education: high school diploma (n=5), two-year degree (n=3), four-year degree (n=4), Master's degree (n=2), and Juris Doctorate (n=1).
Participants reported using a computer for $8.46\pm3.92$ ($M \pm SD$, range: 2.0--16.5) hours per day, and all were experienced using their chosen accessible computer input device.
Participants used their own computer input devices to remotely operate the robotic body surrogate in Atlanta, GA from locations across the United States (Fig. \ref{fig:map}).

The study included seven distinct computer access methods used to operate the robot: trackball (n=4), touchpad (n=3), head tracking with TrackerPro (n=1) and HeadMouse Extreme (n=2) brand devices, standard mouse (n=2), eye-gaze control via Tobii I-15+ (n=1), voice control via Dragon MouseGrid (n=1), and using a touchpad via mouthstick stylus (n=1).
Participants demonstrated a throughput of $2.36 \pm 1.00$ bits/s ($M \pm SD$, range: 0.71 -- 4.58).
Data from this study may be viewed in \nameref{S1_Dataset} and \nameref{S2_Dataset}.

\subsection*{Study 2: Seven-day In-home Evaluation}

Study 2 characterizes the use of a robotic body surrogate by an expert user with profound motor deficits in his home over seven days. 

The study took place from September 23\textsuperscript{rd} through September 29\textsuperscript{th}, 2016.
This n=1 case study was performed with Henry Evans and his wife and primary caregiver, Jane Evans, whose collaboration was instrumental in the design and development of this system.
The Georgia Institute of Technology IRB approved this study under protocol H15170.
We carried out all procedures according to the approved guidelines, and we obtained written informed consent from Henry and Jane Evans, including specific consent to publish identifying information.
The individuals in this manuscript have also given written informed consent (as outlined in PLOS consent form) to publish these case details.

\subsubsection*{The Expert User}

At the time of the study, Henry Evans was 54 years old, with motor deficits resulting from a brain stem stroke on August 29\textsuperscript{th}, 2002.
Henry is mute, can move his head through a limited range of motion, voluntarily contract his left elbow to a limited degree, and contract his left thumb. 
He states that he retains full sensation.
Henry receives three points on the ARAT with his left arm, and zero with his right.
Henry used a head-tracker and a single mouse button to operate the robot in 16 sessions, for a total of 22:30:50 (hh:mm:ss) of use. 

\subsubsection*{System Configuration, Participant Training, and Robot Maintenance}
For this evaluation, we configured the robot software to run automatically on power-up, bringing the robot to a state where Henry could operate it via the web interface.
We provided Henry a link to access the interface on his local network at the beginning of the trial, which he saved as a bookmark. 
Before the experiment, we reviewed the official Willow Garage safety instructions for the PR2 with both Henry and Jane. 
We provided them with both digital and hard copies of an eight-page User Instruction Manual which we wrote for their reference during this study, including the safety instructions, use of the interface, and possible procedures they could follow for error recovery. 
We trained Jane to power up the PR2, to use the PR2's emergency run-stops, and to use the PlayStation 3 joystick to drive the robot.  

Two researchers lived in the participants' home during the trial period and observed all use of the robot for data collection and participant safety.
During the study, researchers plugged and unplugged the robot's power cord as necessary to allow the participant to move the robot about the house and to maintain sufficient battery charge, and engaged the robot's safety stop between sessions of use for additional safety.
Researchers also connected a power cable to the head of the electric shaver whenever the participant successfully grasped the shaver head with the robot, so that he could operate the tool via the web interface, as this requires significant manual dexterity, but could be eliminated with a wireless, battery-powered shaver in the future.
Researchers did not provide further instructions for or assistance with the operation of the robotic body surrogate during the study period.

\subsubsection*{Sessions of Robot Use}
Use of the robot occurred in sessions, where each session consisted of the time from when the participant started to when he stopped the web control interface.
Researchers made themselves available on short notice at any point throughout the week, though the participant preferred to schedule use of the robot in advance, identifying periods during the day when he wished to perform certain tasks. 
This enabled researchers to prepare for data collection in advance, reducing delays or disruptions.
During use sessions, researchers recorded observations and video and monitored safety.
At the end of each session, we conducted a short debriefing, asking the participant to identify the top-level tasks he attempted during the session, and to rate his success, the usefulness of the robot, and the ease of using the robot, for each task.

During sessions, the robot logged all commands issued from the web interface, though not mouse movements or clicks which did not send commands.
The system logged the position of the robot's joints at 4 Hz, and recorded a 540x960 full-color image from the robot's camera at 0.25 Hz. 
Throughout the week, the system logged the calibration status, battery state, run-stop state, any commands issued via the joystick, and all other data reported via the PR2's diagnostics.
Outside of sessions, we suspended formal observation to allow for a more natural and relaxed environment in the home, with the aim of reducing disruption of the participants' typical routines. 

\subsubsection*{Hierarchical Task Analysis}
We identified tasks and subtasks performed during the use sessions using a hierarchical task analysis style breakdown based on direct observation, video recordings, data collected from the robot, and user interviews \cite{Annett2003}.
As a stopping criterion\cite{Felipe2010}, we identified more fine-grained layers of subtasks until the time-stamped data collected on the robot provided the next level of sub-task data (typically at the level of discrete Cartesian movements by either the user or the robot).
For example, we break down labeling of a portion of a feeding task into `scoop yogurt,' `bring yogurt to mouth,' and `eat yogurt' steps.  
We can then further decompose `scoop yogurt' and `bring yogurt to mouth' based upon the individual motion commands sent to the robot, which were automatically recorded and timestamped on the robot.
This typically only required 2-3 subtask levels, and enabled evaluation of the large quantities of automatically collected data according to the higher-level tasks with which they were associated.
We adjusted the task labeling until we reached consensus between both researchers who had observed the trial and the participant.

\section*{Results}

In this section, we present the results of Study 1 and Study 2. Both Study 1 and Study 2 completed successfully without incident. 

\subsection*{Study 1: Remote Evaluation using a Modified Action Research Arm Test and a Simulated Self-Care Task (n=15) }

15 participants completed all aspects of Study 1, meeting our goals as determined by a statistical power analysis. Below, we present the results from the study, organized into an analysis of the novice users' training time and results from participants controlling the robot to perform the modified ARAT and the simulated self-care task.

\subsubsection*{Participant Training Time}
Participants required $54 \pm 16$ min ($M \pm SD$, range: 34--90 min) to complete the guided training portion, and $18 \pm 8$ min ($M \pm SD$, range: 8--33.5 min) to complete the training evaluation task.
Overall, participants operated the robot for $72 \pm 21$ min ($M \pm SD$, range: 44--114 min) before performing the ARAT evaluation.
In post-hoc analyses we found no significant association between either training time or time to complete the evaluation task and either ARAT scores when operating the robot or improvement in ARAT score when using the robot vs. the participant's own body ($n=15, p>.05$ for Pearson's r correlations).

\begin{figure}[h!]
\centering
\includegraphics[width=1.0\textwidth]{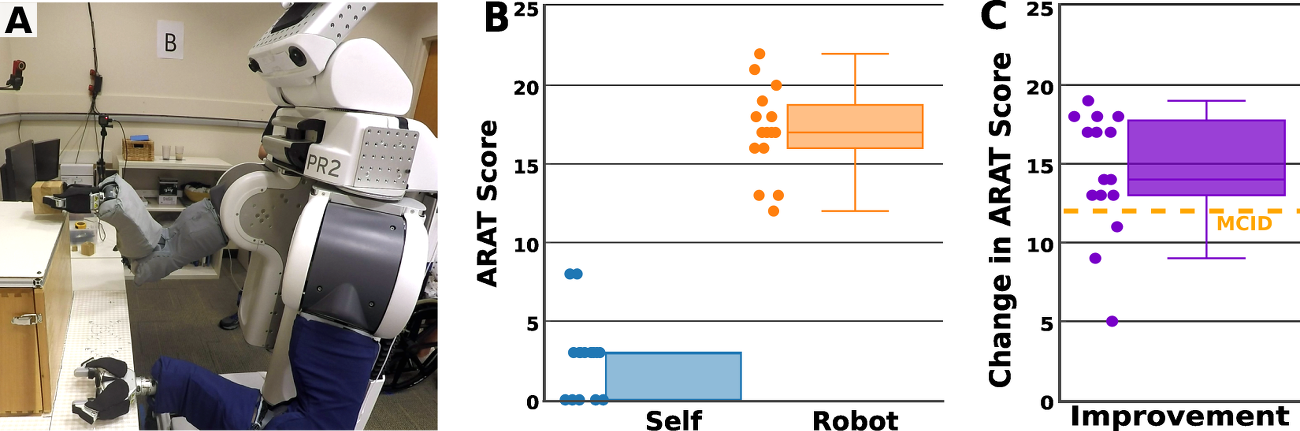}
\caption{{\bf 15 participants with profound motor deficits operated the robotic body surrogate over long distances to perform the Action Research Arm Test (ARAT).} 
(A) A participant remotely performing an item from the ARAT with the robotic body surrogate: grasping, lifting, and placing a 7.5 cm wooden block. 
(B) Comparison of participant ARAT scores without (left) and with (right) the robot ($n=15, W=120, p=0.00035$). 
(C) ARAT score improvements vs. minimal clinically important difference (MCID) reported in literature\cite{Lang2008} (MCID=12, $n=15, W=96, p=0.00147$).}
\label{fig:arat}
\end{figure}

\subsubsection*{Modified ARAT with the Robot}
Participants showed significant improvement on the modified ARAT when remotely operating the robotic body surrogate versus the unassisted remote ARAT assessment (Fig. \ref{fig:arat}). 
Using their own bodies, participants achieved a median score of 3/57 (range: 0--8), with 13/15 participants scoring either 0 or 3 (unable to raise either hand against gravity). Using the robot remotely, participants' ARAT scores averaged 17.07 $\pm$ 2.87 (M $\pm$ SD, median: 17, range: 12--22, Fig. \ref{fig:arat}B, \nameref{S2_Video}), a significant improvement ($n=15, W=120, p=0.00035$). Participants experienced difficulty grasping the smallest object (1 cm marble), and performed best on larger objects that fit easily into the robot's gripper (e.g. 5 cm wooden block). 
The score improvement of participants in our study when using the robot also significantly exceeded a conservative estimate (12 points) of the minimal clinically important difference (MCID) ($n=15, W=96, p=0.00147$, Fig. \ref{fig:arat}C). 
Van der Lee et al.\cite{vanderLee2001reliability} and Lang et al.\cite{Lang2008} report the MCID for the ARAT as 5.7 and 12 points, respectively.

\begin{figure}[h!]
\centering
\includegraphics[width=0.85\textwidth]{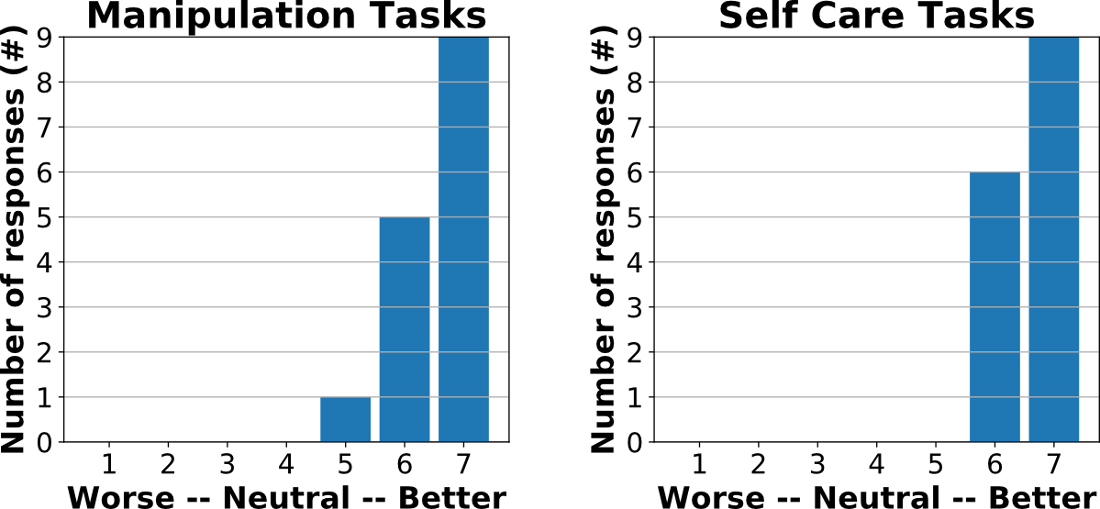}
\caption{{\bf Participants indicated that the robotic body surrogate would provide a significantly meaningful improvement in their ability to perform both manipulation tasks (n=15, W=105, p=0.00036) and self-care tasks (n=15, W=120, p=0.00024).}
Participants were asked to complete the sentence ``using the robotic system rather than my own arms would make my ability to perform [manipulation tasks / self-care tasks]..." using a seven-point scale, with possible responses (based on \cite{Lang2008}:  1. Much worse, 2. Meaningfully worse, 3. A little worse, but not meaningfully, 4. Neither better nor worse, 5. A little better, but not meaningfully, 6. Meaningfully better, and 7. Much better.
The charts shows the distribution of responses to each form of this question.
Significance was evaluated using a 1-tailed, 1-sample Wilcoxon signed rank test vs. a rating of 5--`A little better, but not meaningfully.'}
\label{fig:meaningfulness}
\end{figure}

\begin{figure}[h!]
\centering
\includegraphics[width=0.82\textwidth]{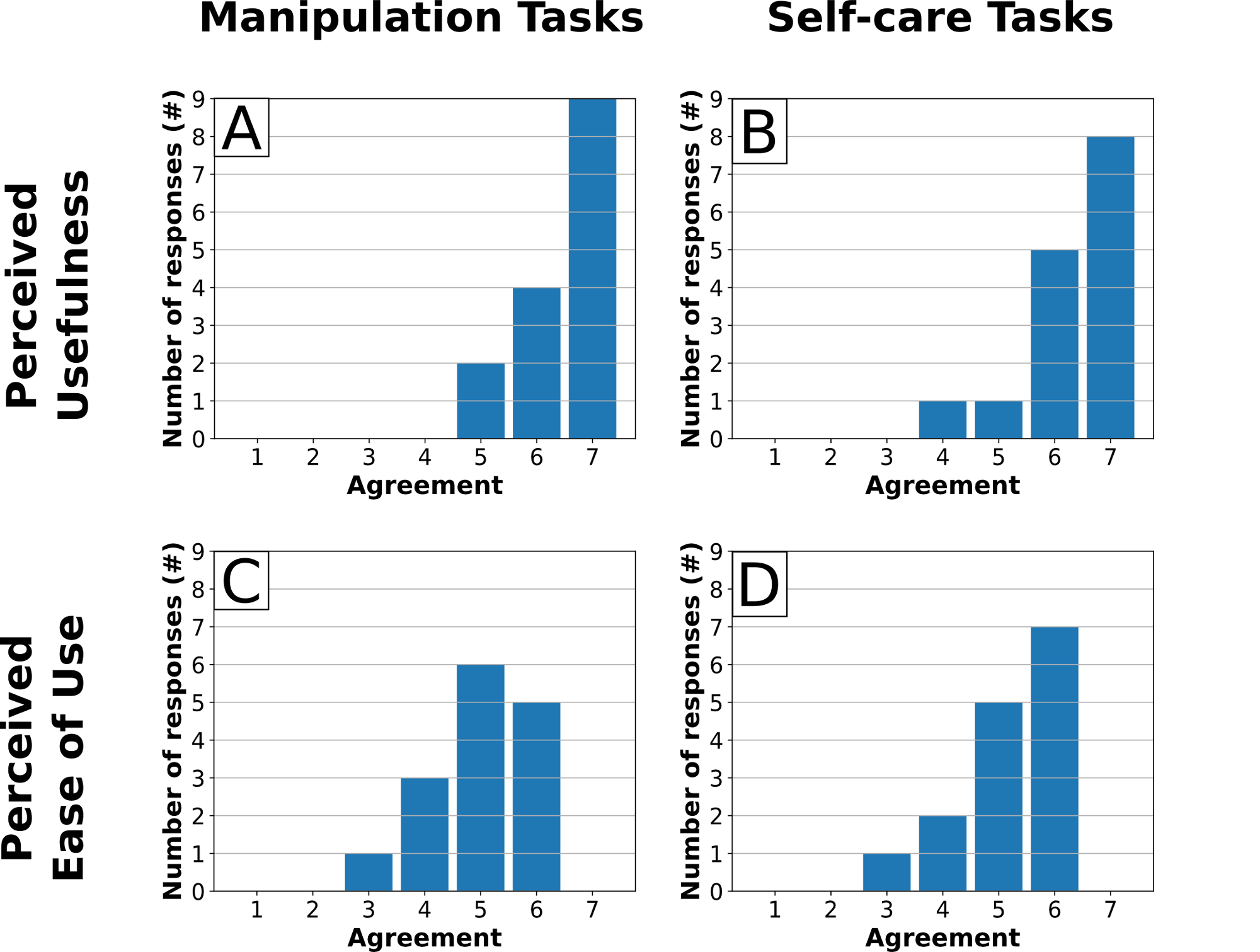}
\caption{\textbf{Participants significantly agreed that the system was both useful (Use) and easy to use (Ease) for both manipulation tasks (Manip.) and self-care tasks (Self).}
(A) Use-Manip.: W=120, p=0.00026. 
(B) Use-Self: W=105, p=0.0004. 
(C) Ease-Manip.: W=74, p=0.0026. 
(D) Ease-Self: W=87.5, p=0.0014. 
$n=15$ for all.
}
Participants were asked to rate their agreement with the statements ``the robotic system is (easy to use / useful) for performing (manipulation/self care) tasks'' using a seven-point scale.
Allowed responses were: 1. Strongly disagree, 2. Disagree, 3. Somewhat disagree, 4. Neither agree nor disagree, 5. Somewhat agree, 6. Agree, and 7. Strongly agree.
Charts show the distribution of responses to each combination of (useful/easy to use) and (manipulation tasks/self care tasks).
Significant was evaluated using a 1-tailed, 1-sample Wilcoxon signed rank test vs. a rating of 4--`Neither agree nor disagree.'
\label{fig:tam_results}
\end{figure}

Participants indicated that the body surrogate would provide a significantly meaningful improvement in ability to perform manipulation tasks ($n=15, W=105, p=0.00036$) and self-care tasks ($n=15, W=120, p=0.00024$) as quantified by responses to a seven-point rating scale based on Lang's questionnaire\cite{Lang2008}, significantly exceeding a rating of `better, but not meaningfully' in both cases (Fig. \ref{fig:meaningfulness}).
Participants also significantly agreed that the system was both useful and easy to use for both manipulation and self-care tasks ($n=15, p<0.003$ in all four cases, see Fig. \ref{fig:tam_results}). 
Participants did not indicate a preference for the robot over a human caregiver for either manipulation tasks ($n=15, W=30.5, p=0.856$) or self-care tasks ($n=15, W=51, p=0.719$).
For both types of task, the median response was ‘neither agree nor disagree,’ and the mode response was ‘agree.’

Task completion when using our system was slow compared to able-bodied performance. 
Participants averaged 4:29 $\pm$ 1:55 (m:ss, $M\pm SD$) for each completed ARAT item, while able-bodied performance without a robot is $\approx$ five seconds or less. 
For individuals with profound motor deficits, slow task performance would still increase independence by enabling people to perform tasks for themselves that would not be possible without assistance.

In post-hoc analyses, we found no significant relationship between any demographic or experience data and either ARAT scores when operating the robot or improvement in ARAT score when using the robot vs. the participant's own body ($n=15, p>.05$ for Pearson's r correlations).
Additionally, we found no significant effects between source of motor deficit, the type of pointing device used, or throughput with the chosen pointing device and either ARAT scores when operating the robot or improvement in ARAT score when using the robot vs. the participant's own body ($n=15, p>.05$ for Kruskal-Wallis H tests).

\begin{figure}[h!]
\centering
\includegraphics[width=1.0\textwidth]{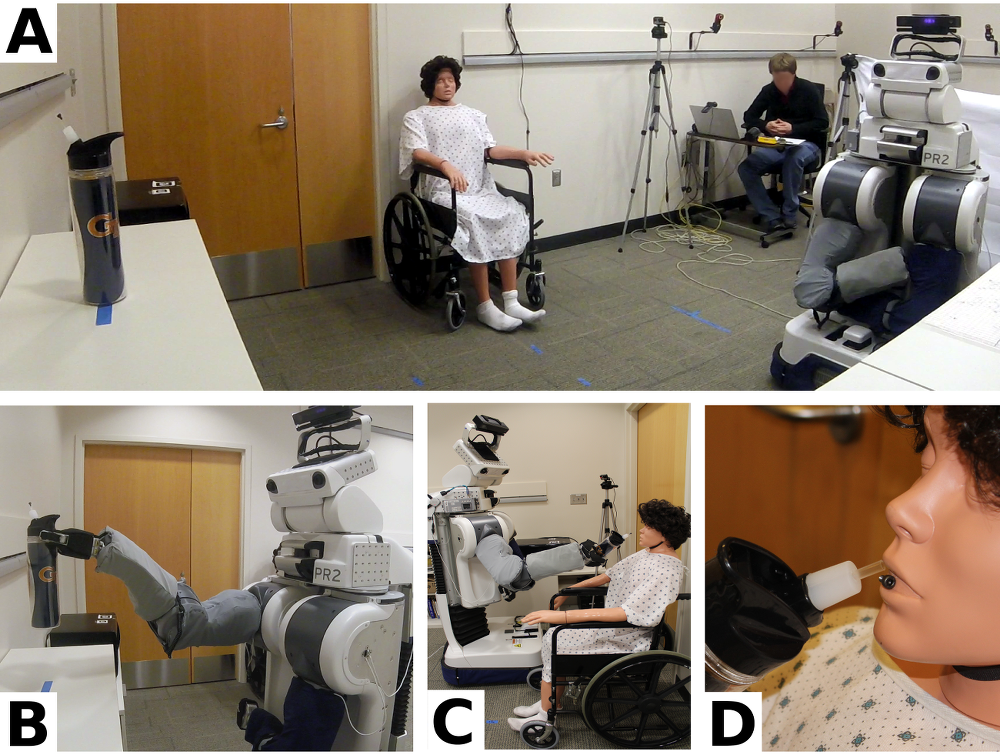}
\caption{{\bf 15 participants with profound motor deficits operated the robotic body surrogate over long distances to simulate getting themselves a drink.} 
(A) The layout of the task room at the beginning of the tasks.  The bottle (left) is placed on a shelf, approximately two meters in front of the robot, and the mannequin in a wheelchair is placed nearby.  The observing researcher sits in the back of the room.
(B) A participant remotely retrieving the water bottle.
(C) A participant reaching and rotating the grasped bottle toward the mannequin's mouth.
(D) The straw in the bottle at the center of the mannequin's mouth, showing the small screw adhered to the magnet behind the mannequin's mouth, indicating successful completion of the task.}
\label{fig:drinking_task}
\end{figure}

\subsubsection*{Simulated Self-care Task with the Robot}
12 of the 15 participants (80\%) successfully completed the simulated self-care task of getting a drink and bringing it to the mouth of a nearby mannequin (Fig. \ref{fig:drinking_task}).
Of the three who did not complete the task successfully, two grasped the bottle, but failed to bring the tip of the straw to the mannequin's mouth with sufficient accuracy before giving up.
The third was unable to successfully grasp the water bottle from the shelf before giving up.
A researcher then placed the bottle into the robot's gripper, at which point the participant successfully brought the tip of the straw to the mannequin's mouth.

When successful (14/15, 93.3\% of participants), participants grasped and lifted the bottle from the shelf in 647s $\pm$ 589s (M $\pm$ SD, range: 195s--2329s).
When successful (13/15, 86.6\% of participants), participants brought the straw tip to the mannequin's mouth with sufficient accuracy to secure the magnet in 1194s $\pm$ 1491s (M $\pm$ SD, range: 217s--4407s) after first grasping and lifting the bottle.
Participants who completed the full task without assistance (12/15, 80\% of participants) completed the task in 1715s $\pm$ 1502s (M $\pm$ SD, range: 465s--4602s). 

\subsection*{Study 2: Seven-day In-home Evaluation}

We present notable results from Study 2 in this section. We have also included videos and data from the study that can provide additional insight. 

\subsubsection*{Hierarchical Task Analysis}

\begin{figure}[h!]
\centering
\includegraphics[width=0.85\textwidth]{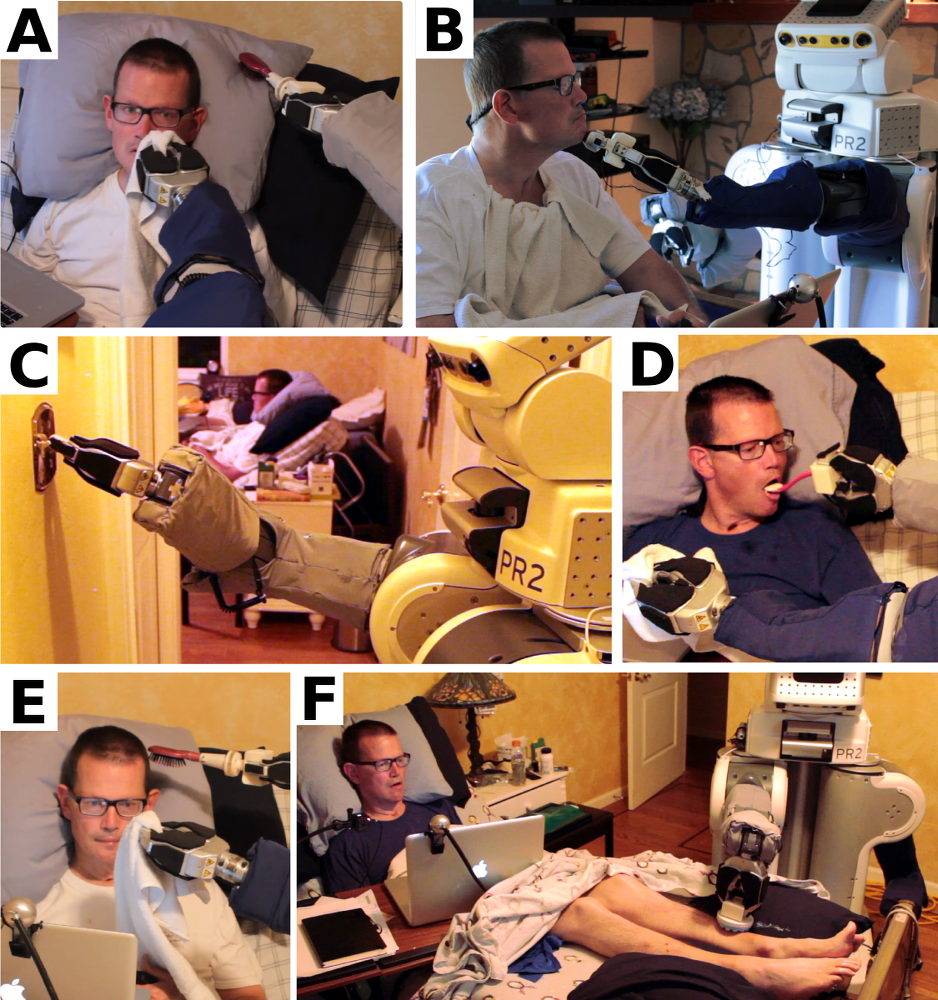}
\caption{{\bf Henry Evans performed 59 separate tasks, including ten distinct types of self-care task and seven distinct types of household task, during the seven-day in-home evaluation.} 
This figure shows Henry Evans performing a selection of these tasks, including:
(A) Wiping his face.
(B) Shaving his face.
(C) Flipping a light switch (Henry visible in background).
(D) Feeding himself yogurt.
(E) Scratching his head.
(F) Applying lotion to his legs.}
\label{fig:henry}
\end{figure}

Following the seven-day, in-home evaluation, our hierarchical task analysis identified 59 `top-level' tasks performed by Henry during the study.
`Top-level' tasks are not subtasks of any other ongoing tasks, but may themselves be composed of one or more sub-tasks.
We identified ten types of self-care tasks and seven types of household tasks that Henry performed, many of which Henry performed multiple times during the week, with each instance counting separately toward the 59 total tasks (Fig. \ref{fig:henry}).
The self-care tasks included feeding himself yogurt (\nameref{S3_Video}), wiping his mouth, scratching his head, applying lotion to his legs (\nameref{S4_Video}), and shaving part of his face.  Henry used the robot both from his bed and his wheelchair.  He operated the robot from his bed for 85.9\% of the total use time, but preferred to shave himself while seated in his wheelchair, as this allowed the robot to reach both sides of his face. 

Detailed hierarchical task data with associated timing information from the seven-day deployment can be found in \nameref{S3_Dataset}. 

\subsubsection*{Human Preparation of the Environment}
We provided a number of common items, including an electric toothbrush, electric razor, hairbrush, lotion applicator, and silicone spoon adapted with handles designed for secure grasping by the robot's gripper.
Henry requested assistance with preparing the environment for a number of his tasks. He typically asked for tools or other items to be set out in a nearby location where he could grasp them with the robot, such as placing a spoon and a bowl of yogurt beside his bed. 
This is comparable to the role a caregiver could play.
Henry then controlled the robot to collect the items and perform the intended task, including navigating around his home, grasping the items, and positioning the robot appropriately for performing the selected task.

\subsubsection*{Examples of Anticipated Tasks}
Although task performance was slow and of variable quality, Henry was able to perform a variety of tasks of his choosing that he would not have been able to perform without assistance. 
When feeding himself yogurt, Henry was able to achieve approximately one bite every five minutes after practice.
When shaving his face, Henry was able to shave his cheeks and jaw effectively, but found it difficult to properly orient the tool to shave his neck and under his chin, and caused some skin irritation in attempting to do so.
Henry was able to wipe his mouth effectively, and often did so between bites of yogurt when feeding himself.
When applying lotion to his leg, Henry used the robot to remove a blanket covering his legs, and was able to apply lotion to parts of his shins, though he indicated that the lotion applicator tool did not perform as well as he would have liked, stating that ``the robot work[ed] better than [the lotion applicator].''

\subsubsection*{A Novel Task}
% TOPIC: Results from 7-day deployment
Henry also discovered an unanticipated use for the robot. He controlled the robot to simultaneously hold out a hairbrush to scratch his head and a towel to wipe his mouth (Fig. \ref{fig:henry}A,E, \nameref{S5_Video}). This allowed him to remain comfortable for extended periods of time in bed without requesting human assistance (two sessions approximately $2.5$ hours and $1$ hour in length). Henry stated that ``it completely obviated the need for a human caregiver once the robot was turned on (always the goal)," and that ``once set up, it worked well for hours and kept me comfortable for hours.''
This was a task which designers had not anticipated, and was the most successful use of the robot in terms of task performance and user satisfaction, as the deployed research system provided a clear, consistent benefit to the user and reduced the need for caregiver assistance during these times.
Although trained to do so, Jane did not directly operate the robot using the joystick during the week.

\section*{Discussion}

In this section, we discuss the results and implications of the research.

\subsection*{Recruiting More Broadly via Remote Participation}

By using broadly accessible technologies that allow for remote operation, a comparatively large number of individuals with profound motor deficits were able to participate in our studies.
This not only strengthens our results, but also demonstrates how these methods can enable improvements in future research by more effectively reaching a traditionally underserved population. The participants in our studies have profound motor deficits, comparable to those of participants in studies of highly-invasive BCI's. 
Because of the difficulty of recruiting participants from this population, many prior studies have had only one or a few participant(s) \cite{Hochberg2006,Hochberg2012,Pandarinath2017,Collinger2013,Ajiboye2017}.
Similarly, previous studies in assistive robotics have often had few participants and/or participants with less profound motor deficits \cite{Tsui2015, Wang13, Soekadar2016, Bien04}, and in some cases include unimpaired individuals to evaluate assistive technologies \cite{Jain2015, Meng2016, Gopinath2017}.

\subsection*{The Modified ARAT}

Our use of a standardized evaluation from the medical community has a number of benefits. In administering the modified ARAT, we closely follow \cite{Yozbatiran2008} so that the reported results will be meaningful with respect to ARAT results from other contexts, and interpretable to others familiar with the test, especially clinicians. This also allows the same test to be used in future robotic studies in a way that can reflect performance improvements enabled through alternative software and/or hardware by comparison with the performance data presented here. Furthermore, normative data for ARAT performance provides a reference for comparing our results with able-bodied performance of the same task. 

Application of our modified ARAT assessment needs to be performed carefully in order for the results to be pertinent to real-world use of a robotic system. For example, an autonomous system specialized for the specific objects and tasks in the ARAT might result in a favorable score, yet perform poorly during real-world use with common objects and tasks.  We mitigated this risk by only incorporating low-level robot autonomy, coupling our ARAT study with a study in a real home, and being careful not to design the interface specifically for the ARAT. 

\subsection*{Depth Perception}

Even with the provided ``3D Peek'' view, the lack of effective depth perception with this system presents a challenge to remote operation, especially when manipulating small objects. 3D displays may be beneficial, but the required hardware and software is much less common than for conventional flat panel displays, though this may change with time. For example, consumer virtual reality systems are becoming more common.  

Interestingly, Henry Evans spent 80\% of his time operating the robot with a direct line of sight to the robot. Self-care tasks inherently require that the user be co-located with the robot, which reduces 3D perception issues. It is unclear if Henry would have used the system more frequently out of his line of sight if 3D perception issues were improved. 

\subsection*{Time to Complete Tasks}

The time required for task completion with the system exceeded able-bodied performance by a wide margin, to the extent that adaptation of the ARAT time limits was a necessary modification. This is due to multiple factors, including the slow speed of the robot, the use of discrete Cartesian movements to perform complex manipulations, and the tendency to perform actions carefully, as errors such as knocking over an object are often difficult to correct. Despite this, participant perceptions of usefulness and ease of use were both significantly positive. Perceived ease of use and perceived usefulness have been found to be predictive of acceptance of technologies\cite{TAM_Davis89, TAM2}, including robots\cite{Ezer2009}. This supports the idea that even limited functionality can have a meaningful impact for individuals with profound motor deficits. It also presents opportunities for future research into methods that might improve operation speed.

Due to the slow speed of the robot and the complexity of operation, participants may not have been significantly limited in their performance by their assistive pointing device, but instead by the operation of the system. This may account for the absence of detected effects between participants' performance using the robot and their own level of impairment, demographics, daily computer use, choice of pointing device, or Fitts's law throughput. If the system operated more quickly, effects of chosen computer interfaces might become evident. 

\subsection*{Task Performance and Opportunities for Improvement}

Task performance presents a notable limitation of this system that merits further investigation, as highlighted by Henry's variable success across tasks.
Based on our hierarchical task analysis, we found that the expert user spent considerable time performing tool acquisition and other ancillary tasks, such as driving the robot to and from the location in the home where it remained between sessions of use.
In such cases, limited semi-autonomous capabilities, such as automatic tool-grasping or autonomous navigation, might reduce task completion time. Task-specific automation might also assist with performance quality, such as giving the robot the ability to follow surface contours automatically, which could improve performance on tasks such as shaving or applying lotion. Similarly, as shown in prior work, task-specific applications for common tasks such as feeding assistance and shaving assistance can utilize specialized interfaces and task-specific robot autonomy \cite{Hawkins2014,Park2017}. 

Ultimately, a combination of approaches may prove beneficial. For example, a robotic body surrogate could provide task-specific automation for common tasks and an interface similar to what we presented for uncommon and novel tasks. In general, methods to reduce the rate of errors and increase operational speed merit further investigation.

%, including autonomy, BCIs, and combinations of approaches. 

\subsection*{Implications for Robotic Body Surrogates}

By using the ARAT, we showed that our system provides improvements exceeding minimal clinically important difference (MCID) as established by \cite{Lang2008}. This indicates that our system could produce a meaningful improvement in the ability of members of our target population to perform everyday and self-care tasks independently through the use of a robotic body surrogate.

Because of the constrained nature of the ARAT, we also asked participants to perform a more complex simulated self-care task. 
As this task required the coordinated use of more of the robot's degrees of freedom than the ARAT, the general success of the novice participants suggests that the system can be used for more complex, real-life tasks requiring both mobility and manipulation.

The ability of remote participants, who never saw the robot in person, to use the system effectively despite this limited training and experience supports the system's ease of use. The training and experience in the study are especially limited compared with the potential experience an individual might gain using this type of technology if available long-term in an in-home setting. It seems likely that users' understanding and comfort with the system, and ultimately performance, would improve with additional experience.

While Study 1 shows that a variety of users with profound motor deficits were able to operate the robotic body surrogate effectively, it does so only in a limited research setting, and for highly constrained tasks. In contrast, the Study 2 shows that an experienced user with profound motor deficits can operate the robotic body surrogate effectively in-person, for a wide variety of tasks, and in a real-world setting. Henry's use patterns and choice of tasks reflect his particular needs, preferences, and lifestyle, and were influenced by both his daily routine and his caregivers. Within this context, Henry used the robot for a variety of tasks appropriate to his needs and outside of the scope of those tested in Study 1. He was also able to identify new opportunities for use. 

The specific robotic hardware we used in our study is not sufficiently robust for longer-term use, nor is it cost-effective for consumers. Despite this, our results support the value of this type of robotic body surrogate for increasing user independence and meeting a variety of user needs, including needs unanticipated by designers. 

Our results suggest that this assistive technology can be made widely accessible. The participants in both studies have profound, bilateral motor impairments, with most being unable to lift either arm against gravity. We have shown that members of this target population can operate the complex robotic device effectively using off-the-shelf assistive input devices. We would expect diverse users with lesser motor impairments to also be able to effectively use the system, although users with lesser impairments may gain less benefit from it. 

\section*{Conclusion}
Overall, we have shown that people with profound motor deficits can effectively use robotic body surrogates at home and in remote locations. The participants in our study had a variety of impairments and used a web browser with their preferred off-the-shelf assistive input devices, which suggests that this type of assistive technology could be used by a diverse range of people.  One participant also operated the robotic body surrogate in a home setting over a seven day period with only limited assistance, indicating that this assistive technology can operate effectively outside the context of a laboratory evaluation.  Together, these results suggest that robotic body surrogates could provide improved independence and self-care self-efficacy for individuals with profound motor deficits.

\section*{Acknowledgments}
We thank Henry and Jane Evans for their numerous contributions to our research. We thank Henry Clever and Newton Chan for assistance in conducting the experiments. We thank Vincent Dureau for his feedback on the interface. We also thank Dr. Lena Ting, Dr. Karen Feigh, Dr. Randy Trumbower, and Dr. Chethan Pandarinath for their comments on early drafts of this manuscript.

\section*{Supporting information}

\paragraph*{S1 Video}
\label{S1_Video}
{\bf The Robotic Body Surrogate and Augmented Reality Interface.} The robotic body surrogate, along with the augmented reality (AR) interface used to operate it via a web-browser.  The video demonstrates all control modes and sensor displays present in the interface.

\paragraph*{S2 Video}
\label{S2_Video}
{\bf A User with Profound Motor Deficits Earns 22 Points on the ARAT during Long-distance Operation of the Robotic Body Surrogate.} A remote user operating the robotic body surrogate to earn all 22 points on the ARAT expected to be possible when using the robot remotely. The observing researcher is visible in the background of the video. 

\paragraph*{S3 Video}
\label{S3_Video}
{\bf Henry Evans Uses the Robotic Body Surrogate to Feed Himself Yogurt.} Henry Evans, lying in bed, operates the PR2 to feed himself a scoop of yogurt. Henry grasped the spoon and scooped the yogurt from the bowl (not shown).

\paragraph{S4 Video}
\label{S4_Video}
{\bf Henry Evans Uses the Robotic Body Surrogate to Apply Lotion to His Legs.} Henry Evans, lying in bed, controls the PR2 in his own home to grasp a lotion applicator, move a blanket from atop his legs, apply lotion to his shins, replace the lotion applicator, and park the robot in an out-of-the-way location.

\paragraph*{S5 Video}
\label{S5_Video}
{\bf Henry Evans Uses the Robotic Body Surrogate to Wipe His Mouth and Scratch His Head.} Henry Evans, lying in bed, operates the PR2 to wipe his face with a towel and scratch his head with a hairbrush. Henry conceived of this method of assistance, which kept him comfortable for extended periods without caregiver assistance.

\paragraph*{S1 Dataset}
\label{S1_Dataset}
{\bf Participant data from the ARAT evaluation} These are the data from the evaluation of the system with 15 users with profound motor deficits.  This includes demographic, impairment, and computer access data, data from training sessions, and the reported results for the ARAT, simulated self-care task, and the associated seven-point rating scale responses.

\paragraph*{S2 Dataset}
\label{S2_Dataset}
{\bf ARAT item data from the ARAT evaluation} These are the score and timing data from the evaluation of the system with 15 users with profound motor deficits completing the Action Research Arm Test by operating the robotic body surrogate remotely.

\paragraph*{S3 Dataset}
\label{S3_Dataset}
{\bf Data from the seven-day in-home evaluation} These are the identified tasks and associated timing information recorded during the seven-day deployment of the system in the home of Henry Evans.

\paragraph*{S1 Source Code}
\label{source_code}
{\bf Source code for the robotic system and interface} This includes all custom code used in the described system and related evaluations.

\end{document}